\documentclass[10pt,twocolumn,letterpaper]{article}

\usepackage{booktabs}

\usepackage{cvpr}
\usepackage[utf8]{inputenc} 

\usepackage{times}
\usepackage{epsfig}
\usepackage{graphicx}
\usepackage{amsmath}
\usepackage{amssymb}
\usepackage{subfigure}
\usepackage[page]{appendix}

\usepackage[T1]{fontenc}
\usepackage{CJKutf8}
\usepackage[english]{babel}

\usepackage[breaklinks=true,bookmarks=false,colorlinks=true,linkcolor=red,citecolor=blue]{hyperref}
\cvprfinalcopy 


\makeatletter
\g@addto@macro{\UrlBreaks}{\UrlOrds}
\makeatother

\newcommand{\subtab}[1]{\medskip\noindent\textbf{#1}\;}

\begin{document}
\title{Compounding the Performance Improvements of Assembled Techniques\\in a Convolutional Neural Network} 

\author{
Jungkyu Lee, Taeryun Won, Tae Kwan Lee, Hyemin Lee, Geonmo Gu, Kiho Hong\thanks{Corresponding author.}
\\
\\Clova Vision, NAVER Corp.
\\{\tt\small \{jungkyu.lee, lory.tail, taekwan.lee, hmin.lee, geonmo.gu, kiho.hong\}@navercorp.com}}

\maketitle

\begin{abstract}
Recent studies in image classification have demonstrated a variety of techniques for improving the performance of Convolutional Neural Networks (CNNs). 
However, attempts to combine existing techniques to create a practical model are still uncommon. 
In this study, we carry out extensive experiments to validate that carefully assembling these techniques and applying them to basic CNN models (\eg, ResNet and MobileNet) can improve the accuracy and robustness of the models while minimizing the loss of throughput.
Our proposed assembled ResNet-50 shows improvements in top-1 accuracy from 76.3\% to 82.78\%, mCE from 76.0\% to 48.9\% and mFR from 57.7\% to 32.3\% on ILSVRC2012 validation set.
With these improvements, inference throughput only decreases from 536 to 312.
To verify the performance improvement in transfer learning, fine grained classification and image retrieval tasks were tested on several public datasets and showed that the improvement to backbone network performance boosted transfer learning performance significantly. Our approach achieved 1st place in the iFood Competition Fine-Grained Visual Recognition at CVPR 2019\footnote{https://www.kaggle.com/c/ifood-2019-fgvc6/leaderboard}, and the source code and trained models will be made publicly available\footnote{https://github.com/clovaai/assembled-cnn}.
\end{abstract}

\section{Introduction}
Since the introduction of AlexNet~\cite{krizhevsky2012imagenet}, many studies have mainly focused on designing new network architectures for image classification to increase accuracy.
For example, new architectures such as Inception~\cite{szegedy2015going}, ResNet~\cite{he2016deep}, DenseNet~\cite{huang2017densely}, NASNet~\cite{zoph2018learning}, MNASNet~\cite{tan2019mnasnet} and EfficientNet~\cite{tan2019efficientnet} have been proposed.
Inception introduced new modules into the network with convolution layers of different kernel sizes. 
ResNet utilized the concept of skip connection, and DenseNet added dense feature connections to boost the performance of the model. In addition, in the area of AutoML, network design was automatically decided to create models such as NASNet and MNASNet.
EfficientNet proposes an efficient network by balancing the resolution, height, and width of the network. 
The performance of EfficientNet for ILSVRC2012 top-1 accuracy was greatly improved relative to AlexNet. 

\begin{table*}[t]
\small{
\begin{center}
\begin{tabular}{@{}l|cccc@{}}
\toprule
\multicolumn{1}{c|}{\textbf{Model}} &
\multicolumn{1}{c}{\textbf{Top-1}} &
\multicolumn{1}{c}{\textbf{mCE}} &
\multicolumn{1}{c}{\textbf{mFR}} &
\textbf{\begin{tabular}[c]{@{}c@{}}Throughput\\ \end{tabular}} \\ \midrule
EfficientNet B4~\cite{tan2019efficientnet} + AutoAugment~\cite{cubuk2018autoaugment} & 83.0      & 60.7  & -     & 95  \\
EfficientNet B6~\cite{tan2019efficientnet} + AutoAugment~\cite{cubuk2018autoaugment} & 84.2      & 60.6  & -     & 28  \\
EfficientNet B7~\cite{tan2019efficientnet} + AutoAugment~\cite{cubuk2018autoaugment} & 84.5      & 59.4  & -     & 16  \\ \midrule
ResNet-50~\cite{he2016deep} (baseline)     & 76.3      & 76.0       & 57.7   & 536 \\ \midrule
Assemble-ResNet-50\; (ours)                & 82.8      & 48.9       & 32.3   & 312 \\ 
Assemble-ResNet-152 (ours)                 & 84.2      & 43.3       & 29.3   & 143 \\ \bottomrule
\end{tabular}
\end {center}
}
\caption{
    \textbf{Summary of key results.} \textit{Top-1} is ILSVRC2012 top-1 validation accuracy. 
    \textit{mCE} is mean corruption error and \textit{mFR} is mean flip rate (Lower is better.)~\cite{hendrycks2019benchmarking}.
    The \textit{Throughput} refers to how many images per second the model processes during inference. 
}
\label{tab:summary-model}
\end{table*}

Unlike these studies which focus on designing new network architecture, He~\etal~\cite{he2019bag} proposes different approaches to improve model performance. 
They noted that performance can be improved not only through changes in the model structure, but also through other aspects of network training such as data preprocessing, learning rate decay, and parameter initialization. 
They also demonstrate that these minor ``tricks'' play a major part in boosting model performance when applied in combination. 
As a result of using these tricks, ILSVRC2012 top-1 validation accuracy of ResNet-50 improved from 75.3\% to 79.29\% and MobileNet improved from 69.03\% to 71.90\%.  
This improvement is highly significant because it shows as much performance improvement as a novel network design does.

Inspired by \cite{he2019bag}, we conducted a more extensive and systematic study of \textit{assembling} several CNN-related techniques into a single network. 
We first divided the CNN-related techniques into two categories: network tweaks and regularization. 
Network tweaks are methods that modify the CNN architectures to be more efficient.
(\eg, SENet~\cite{hu2018squeeze}, SKNet~\cite{li2019selective}).
Regularization includes methods that prevent overfitting by increasing the training data through data augmentation processes such as AutoAugment~\cite{cubuk2018autoaugment} and Mixup~\cite{zhang2017mixup}, or by limiting the complexity of the CNN with processes such as Dropout~\cite{srivastava2014dropout}, and DropBlock~\cite{ghiasi2018dropblock}.
We then systematically analyze the process of assembling these two types of techniques through extensive experiments and demonstrate that our approach leads to significant performance improvements.

In addition to top-1 accuracy, mCE, mFR and throughput were used as \textit{performance indicators} for combining these various techniques.
Hendrycks~\etal~\cite{hendrycks2019benchmarking} proposed mCE (mean corruption error) and mFR (mean flip rate). mCE is a measure of network robustness against input image corruption, and mFR is a measure of network stability on perturbations in image sequences.
Moreover, we used throughput (images/sec) instead of the commonly used measurement of FLOPS (floating point operations per second) because we observed that FLOPS is not proportional to the inference speed of the actual GPU device.

\medskip\noindent Our contributions can be summarized as follows:
\begin{enumerate}
\item By organizing the existing CNN-related techniques for image classification, we find techniques that can be assembled into a single CNN.
We then demonstrate that our resulting model surpasses the state-of-the-art models with similar accuracy in terms of mCE, mFR and throughput (Table~\ref{tab:summary-model}).
\item  We provide detailed experimental results for the process of assembling CNN techniques and release the code for accessibility and reproducibility.
\end{enumerate}

\section{Preliminaries}

\label{sec:basemodel}

Before introducing our approach, we describe default experimental settings and evaluation metrics used in Sections \ref{sec:assembling} and \ref{sec:Experiment Results}.

\subsection{Training Procedure}

We use the official TensorFlow~\cite{abadi2016tensorflow} ResNet~\footnote{https://github.com/tensorflow/models} as base code. 
The ILSVRC2012~\cite{russakovsky2015imagenet} dataset is used to train and evaluate models.
All models were trained on a single machine with 8 Nvidia Tesla P40 GPUs compatible with the CUDA 10 platform and cuDNN 7.6. TensorFlow version 1.14.0 was used.

The techniques proposed by He~\etal~\cite{he2019bag} are basically applied to all our models described in Section~\ref{sec:assembling}. 
We briefly describe the default hyperparameters and training techniques as follows.

\subtab{Preprocessing}
In the training phase, a rectangular region is randomly cropped using a randomly sampled aspect ratio from 3/4 to 4/3, and the fraction of cropped area over whole image is randomly chosen from 5\% to 100\%. Then, the cropped region is resized to $224\times224$ and flipped horizontally with a random probability of 0.5 followed by the RGB channel normalization. During validation, shorter dimension of each image is resized to 256 pixels while the aspect ratio is maintained. Next, the image is center-cropped to $224\times224$, and the RGB channels are normalized.

\subtab{Hyperparameter}
We use 1,024 batch size for training which is close to the maximum size that can be received by a single machine with 8 P40 GPUs. 
Stochastic gradient descent with momentum 0.9 is used as the optimizer. The initial learning rate is 0.4 and the weight decay is set to 0.0001. The default number of training epochs is 120, but some techniques require different number of epochs. This is explicitly specified when necessary. 

\subtab{Learning rate warmup}
If the batch size is large, a high learning rate may result in numerical instability.
To prevent this, Goyal~\etal~\cite{goyal2017accurate} proposes a warmup strategy that linearly increases the learning rate from 0 to the initial value. The warm-up period is set to the first 5 epochs.
    
\subtab{Zero $\boldsymbol{\gamma}$}
We initialize $\gamma=0$ for all batch-norm layers that sit at the end of residual blocks. 
Therefore, all the residual blocks only return their shortcut branch results in the early stages of training.
This has the effect of shrinking the entire layer at the initial stage and helps training.  
    
\subtab{{Mixed-precision floating point}}
We use mixed-precision floating point in the training phase because mixed-precision accelerates the overall training speed if the GPU supports it~\cite{micikevicius2017mixed}. However, this does not result in the improvement of top-1 accuracy.
    
\subtab{{Cosine learning rate decay}}
The cosine decay schedule~\cite{loshchilov2016sgdr} reduces the initial learning rate to close to 0 at the end of training by following a cosine curve. 

\subsection{Evaluation Metrics}

The selection of metrics used to measure the performance of the model is important because it indicates the direction in which the model is developed.
We use the following three metrics as key indicators of model performance. 

\subtab{Top-1}
The top-1 is a measure of classification accuracy on the ILSVRC2012~\cite{russakovsky2015imagenet} validation dataset. The validation dataset consists of 50,000 images of 1,000 classes. 

\subtab{Throughput}
Throughput is defined as how many images are processed per second on the GPU device.
We measured inference throughput for an Nvidia P40 1 GPU.  
For comparison with other models, we used FP32 instead of FP16 in our experiments, using a batch size of 64. 

\subtab{mCE and mFR}
The mean corruption error (mCE) and the mean flip rate (mFR) were proposed by Hendrycks~\etal~\cite{hendrycks2019benchmarking} to measure the performance of the classification model on corrupted images and network stability on perturbations in image sequences, respectively.

\section{Assembling CNN}
\label{sec:assembling}

In this section, we introduce various network tweaks and regularization techniques to be assembled, and describe the details of the implementation.
We also perform preliminary experiments to study the effect of different parameter choices.

\subsection{Network Tweaks}
\label{sec:network-tweak}

\begin{figure*}[h]
\begin{center}
\includegraphics[width=1\linewidth]{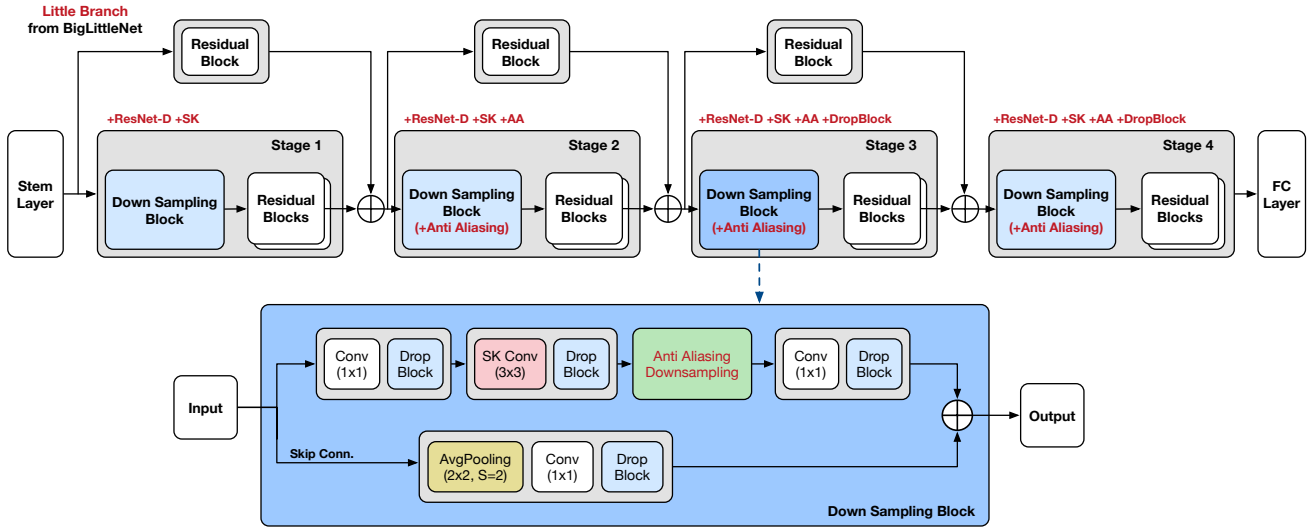}
\end{center}
\caption{
    \textbf{Assembling techniques into ResNet}. We apply network tweaks such as ResNet-D, SK, Anti-alias,  DropBlock, and BigLittleNet to vanilla ResNet.
   In more detail, ResNet-D and SK are applied to all blocks in all stages.
   Downsampling with anti-aliasing is only applied to the downsampling block from Stage 2 to Stage 4.
   DropBlock is applied to all blocks in Stage 3 and Stage 4. 
   Little-Branch from BigLittleNet uses one residual block with smaller width.
   }
\label{fig:architecture}
\end{figure*}

Figure~\ref{fig:architecture} shows the overall flow of our final ResNet model. 
Various network tweaks are applied to vanilla ResNet. The network tweaks we use are as follows.

\subtab{ResNet-D}
ResNet-D is a minor adjustment to the vanilla ResNet network architecture model proposed by He~\etal~\cite{he2019bag}.
It is known to work well in practice and has little impact to computational cost~\cite{he2019bag}.
Three changes are added to the ResNet model.
First, the stride sizes of the first two convolutions in the residual path have been switched. 
Second, a $2\times2$ average pooling layer with a stride of 2 is added before the convolution in the skip connection path.
Last, a large $7\times7$ convolution is replaced with three smaller $3\times3$ convolutions in the stem layer.

\subtab{Channel Attention} 
We examine two tweaks in relation to channel attention.
First, Squeeze and Excitation (SE) network~\cite{hu2018squeeze} focuses on enhancing the representational capacity of the network by modeling channel-wise relationships.
SE eliminates spatial information by global pooling to get channel information only, and then two fully connected layers in this module learn the correlation between channels.
Second, Selective Kernel (SK)~\cite{li2019selective} is inspired by the fact that the receptive sizes of neurons in the human visual cortex are different from each other.
SK unit has multiple branches with different kernel sizes, and all branches are fused using softmax attention.

\begin{figure}[h]
\begin{center}
\includegraphics[width=1.0\linewidth]{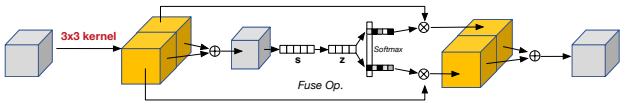}
\end{center}
\caption{\textbf{Modified SK Unit.} We use one $3\times3$ kernel with doubled output channel size instead of $5\times5$ and $3\times3$ kernels.}
\label{fig:modified-sk}
\end{figure}

The original SK generates multiple paths with $3\times3$ and $5\times5$ convolutions, but we instead use two $3\times3$ convolutions to split the given feature map.
This is because two convolutions of the same kernel size can be replaced with one convolution with twice as many channels, thereby lowering the inference cost.
Figure~\ref{fig:modified-sk} shows an SK unit where the original two branches are replaced with one convolution operation.

\begin{table}[h!]
\begin{center}
\scriptsize{
\begin{tabular}{c|l|l|c|cc}
\toprule
\textbf{\begin{tabular}[c]{@{}c@{}}Exp\\No.\end{tabular}} &
\multicolumn{1}{c|}{\textbf{Model}} &
\textbf{\begin{tabular}[c]{@{}c@{}}SK\\Configuration\end{tabular}} &
\textbf{\begin{tabular}[c]{@{}c@{}}SK\\$\boldsymbol{r}$\end{tabular}} &
\multicolumn{1}{c}{\textbf{Top-1}} &
\textbf{Throughput}
\\ \midrule
C0 & R50 (baseline)   & -                & -  & 76.30       & 536  \\
C1 & R50+SE           & -                & -  & 77.40       & 466  \\
C2 & R50+SK           & 3x3+5x5        & 2  & 78.00       & 326  \\
C3 & R50+SK           & 3x3, 2x-channel  & 2  & 77.92       & 382  \\
C4 & R50+SK           & 3x3, 2x-channel  & 16 & 77.57       & 402  \\
C5 & R50+SK+SE        & 3x3, 2x-channel  & 2  & 77.50       & 345  \\ \bottomrule
\end{tabular}
}
\end{center}
\caption{
Result of channel attention with different configurations. R50 is a simple notation for ResNet-50. $r$ is the reduction ratio of SK in the \textit{Fuse} operation.
The piecewise learning rate decay is used in these experiments.
} 
\label{tab:channelattention}
\end{table}

Table~\ref{tab:channelattention} shows the results for different configurations of channel attention.
Compared with SK, SE has higher throughput but lower accuracy (C1 and C2 in Table~\ref{tab:channelattention}). 
Between C3 and C2, the top-1 accuracy only differs by 0.08\% (78.00\% and 77.92\%), but the throughput is significantly different (326 and 382).
Considering this trade-off between accuracy and throughput, we decide to use one $3\times3$ kernel with doubled channel size instead of $3\times3$ and $5\times5$ kernels.
Comparing C3 and C4, we see that changing the setting of reduction ratio $r$ for SK units from 2 to 16 yields a large degradation of top-1 accuracy relative to the improvement of throughput.
Applying both SE and SK (C5) not only decreases accuracy by 0.42\% (from 77.92\% to 77.50\%), but also decreases inference throughput by 37 (from 382 to 345). 
Overall, for a better trade-off between top-1 accuracy and throughput, the configuration of C3 is preferred. 

\subtab{Anti-Alias Downsampling (AA)}
CNN models for image classification are known to be very vulnerable to small amounts of distortion~\cite{xie2019adversarial}.
Zhang~\etal~\cite{zhang2019shiftinvar} proposes AA to improve the shift-equivariance of deep networks.
The max-pooling is commonly viewed as a competing downsampling strategy, and is inherently composed of two operations.
The first operation is to densely evaluate the max operator and second operation is naive subsampling~\cite{zhang2019shiftinvar}. 
AA is proposed as a low-pass filter between them to achieve practical anti-aliasing in any existing strided layer such as strided-conv.
The smoothing factor can be adjusted by changing the blur kernel filter size, where a larger filter size results in increased blur.
In~\cite{zhang2019shiftinvar}, AA is applied to max-pooling, projection-conv, and strided-conv of ResNet.
Table~\ref{tab:aa} shows the experimental results for AA.
We observe that reducing the filter size from 5 to 3 maintains the top-1 accuracy while increasing inference throughput (A1 and A2 in Table~\ref{tab:aa}).
However, removing the AA applied to the projection-conv does not affect the accuracy (A3).
We also observe that applying AA to max-pooling degrades throughput significantly (A1, A2, and A3) compared to A4.
Based on the result, we apply AA only to strided-conv in our model (Green box in Figure~\ref{fig:architecture}).

\begin{table}[h]
\begin{center}
\scriptsize{
\begin{tabular}{c|c|c|c|c|cc}
\toprule
 \textbf{\begin{tabular}[c]{@{}c@{}}Exp\\No.\end{tabular}} &
 \textbf{\begin{tabular}[c]{@{}c@{}}Filter\\Size\end{tabular}} &
 \textbf{\begin{tabular}[c]{@{}c@{}}Max\\Pooling\end{tabular}} &
 \textbf{\begin{tabular}[c]{@{}c@{}}Projection\end{tabular}} &
 \textbf{\begin{tabular}[c]{@{}c@{}}Strided\\Conv\end{tabular}} & \textbf{Top-1} & \textbf{Throughput} \\ \midrule
A0 & -        & X             & X          & X         & 76.30 & 536        \\ \midrule
A1 & 5        & O             & O          & O         & 76.81 & 422        \\
A2 & 3        & O             & O          & O         & 76.83 & 456        \\
A3 & 3        & O             & X          & O         & 76.84 & 483        \\
A4 & 3        & X             & X          & O         & 76.67 & 519        \\ \bottomrule
\end{tabular}
}
\end{center}
\caption{Results for downsampling with anti-aliasing.  
The performance of the model was tested with different configurations for downsampling with anti-aliasing. The piecewise learning rate decay is used in these experiments.
}
\label{tab:aa}
\end{table}

\subtab{Big Little Network (BL)}
BigLittleNet~\cite{chen2018big} applies multiple branches (Big-Branch and Little-Branch) with different resolutions while aiming at reducing computational cost and increasing accuracy. 
The Big-Branch has the same structure as the baseline model and operates at a low image resolution, whereas the Little-Branch reduces the convolutional layers and operates at same image resolution as the baseline model.
BigLittleNet has two hyperparameters, $\alpha$ and $\beta$, which adjust the width and depth of the Little-Branch, respectively. 
We use $\alpha=2$ and $\beta=4$ for ResNet-50 and use $\alpha=1$ and $\beta=2$ for ResNet-152.
The upper small branch in Figure~\ref{fig:architecture} represents the Little-Branch. 
The Little-Branch has one residual block and is smaller in width than the main Big-Branch.
Since BigLittleNet saves budget in computation, the models can be evaluated with a larger input image scale for better performance while maintaining similar throughput ~\cite{chen2018big}.

\subsection{Regularization}
\label{sec:regularization}

\subtab{AutoAugment (Autoaug)}
AutoAugment~\cite{cubuk2018autoaugment} is a data augmentation procedure which learns augmentation strategies from data.
It uses reinforcement learning to select a sequence of image augmentation operations with the best accuracy by searching a discrete search space of their probability of application and magnitude.
We borrow the augmentation policy of Autoaug on ILSVRC2012~\footnote{https://github.com/tensorflow/models/tree/master/research/autoaugment}.

\subtab{Label Smoothing (LS)} In the classification problem, class labels are expressed as one hot encoding. If CNN is trained to minimize cross entropy with this one hot encoding target, the logits of the last fully connected layer of CNN grow to infinity, which leads to over-fitting~\cite{he2019bag}.
Label smoothing~\cite{pereyra2017regularizing} suppresses infinite output and prevents over-fitting. 
We set the label smoothing factor $\epsilon$ to 0.1.

\subtab{Mixup} Mixup~\cite{zhang2017mixup} creates one example by interpolating two examples of the training set for data augmentation. 
Neural networks are known to memorize training data rather than generalize from the data~\cite{zhang2016understanding}.
As a result, the neural network produces unexpected outputs when it encounters data which are different from the distribution of the training set. Mixup mitigates the problem by showing the neural network interpolated examples, and this helps to fill up the empty feature space of the training dataset. 
\begin{table}[h]
\begin{center}
\small{
\begin{tabular}{l|l|c}
\toprule
\multicolumn{1}{c|}{\textbf{Model}} & \multicolumn{1}{c|}{\textbf{Configuration}} & \textbf{Top-1} \\ \midrule
R50D          & LS & 77.37          \\
R50D          & LS + Mixup (type2)     & 78.85      \\
R50D          & LS + Mixup (type1)     & \textbf{79.10}          \\ \bottomrule
\end{tabular}
}
\end{center}
\caption{Result of different Mixup implementation types.}
\label{tab:mixup}
\end{table}

Mixup has two types of implementation. The first type uses two mini batches to create a mixed mini batch. this type of implementation is suggested in the original paper~\cite{zhang2017mixup}. 
The second type uses a single mini batch to create the mixed mini batch by mixing the single mini batch with a shuffled clone of itself. 
The second type of implementation uses less CPU resources because only one mini batch needs to be preprocessed to create one mixed mini batch. 
However, experiments show that the second type of implementation reduces top-1 accuracy (Table \ref{tab:mixup}).
Therefore, in later experiments, we use the first type of implementation. We set the Mixup hyperparameter $\alpha$ to 0.2.

\subtab{DropBlock} 
Dropout~\cite{srivastava2014dropout} is a popular technique for regularizing deep neural networks. 
It prevents the network from being over-fitted to the training set by dropping neurons at random.
However, Dropout does not work well for extremely deep networks such as ResNet~\cite{ghiasi2018dropblock}.
DropBlock~\cite{ghiasi2018dropblock} can remove specific semantic information by dropping a continuous region of activation. Thus, it is efficient for the regularization of very deep networks. 
We borrow the same DropBlock setting used in the original paper~\cite{ghiasi2018dropblock}.
We apply DropBlock to Stage 3 and 4 of ResNet-50 and linearly decay the \(keep\_prob\) hyperparameter from 1.0 to 0.9 during training.

\subtab{Knowledge Distillation (KD)}
Knowledge Distillation~\cite{hinton2015distilling} is a technique for transferring knowledge from one neural network (teacher) to another (student). 
Teacher models are often complex with high accuracy, and a weak but light student model can improve its own accuracy by mimicking a teacher model.
The $T$ hyperparameter of KD was said to be optimal when set to \(2\) or \(3\) in the original paper~\cite{hinton2015distilling}, but we use $T$=$1$ for our model.
Because our model uses Mixup and KD techniques together, the teacher network should also be applied to Mixup. This leads to better performance at lower temperatures because the teacher's signal itself is already smoothed by the Mixup (Table~\ref{tab:kd}). We used EfficientNet B7 as a teacher with 84.5\% of ILSVRC2012 validation top-1 accuracy.
In addition, the logits of the teacher were not computed during the training time, but computed offline before training. The saved teacher logits were then used during training. Although this offline implementation of KD cannot calculate the teacher logits of augmented data (e.g. AutoAugment) during training time, it worked well in our experiments.

\begin{table}[h]
\small{
\begin{center}
\begin{tabular}{@{}l|l|c@{}}
\toprule
\multicolumn{1}{c|}{\textbf{Model}} & \multicolumn{1}{c|}{\textbf{Configuration}} & \textbf{Top-1} \\ \midrule
R50D+SK          & LS+Mixup+DropBlock & 81.40          \\
R50D+SK          & LS+Mixup+DropBlock+KD ($T$=2)     & 81.47          \\
R50D+SK          & LS+Mixup+DropBlock+KD ($T$=1.5)   & 81.50          \\
R50D+SK          & LS+Mixup+DropBlock+KD ($T$=1)     & \textbf{81.69}          \\ \bottomrule
\end{tabular}
\end{center}
}
\caption{Result of the change of KD temperature. We apply KD by varying the temperature $T$ to find the optimal $T$ value. We choose $T=1$ for next experiments.}
\label{tab:kd}
\end{table}

\section{Experiment Results}
\label{sec:Experiment Results}

\subsection{Ablation Study on ResNet}
\label{sec:network-tweaks-result}

In this section we will describe ablation study for assembling the individual network tweaks covered in Section \ref{sec:network-tweak} to find a better model.
The results are shown in Table~\ref{tab:model-tweak}.

\begin{table}[h!]
\small{
\begin{center}
\begin{tabular}{@{}c|l|c|cc@{}} 
\toprule
\textbf{\begin{tabular}[c]{@{}c@{}}Exp.\\No.\end{tabular}} &
\multicolumn{1}{c|}{\textbf{Model}} &
\textbf{\begin{tabular}[c]{@{}c@{}}Input\\Size\end{tabular}} &
\textbf{Top-1} &
\textbf{\begin{tabular}[c]{@{}c@{}}Throughput \end{tabular}}
\\ \midrule
T0  & R50 (baseline)            & 224               & 76.87          & 536    \\ \midrule
T1  & R50\textbf{D}             & 224           & 77.37          & 493    \\
T2  & R50D+\textbf{SK}          & 224           & 78.83          & 359    \\
T3  & R50D+SK+\textbf{BL}       & 224           & 78.26          & 445    \\
T4  & R50D+SK+BL       & \textbf{256}  & 79.27          & 359    \\
T5  & R50D+SK+BL+\textbf{AA}    & 256           & 79.39          & 312    \\ \bottomrule
\end{tabular}
\end{center}
}
\caption{Performance comparison of stacking network tweaks. By stacking the ResNet-D, Selective Kernel (SK), BigLittleNet (BL) and downsampling with anti-aliasing (AA), we have steadily improved the ResNet-50 model with some inference throughput losses. The focus of each experiment is highlighted in boldface.}
\label{tab:model-tweak}
\end{table}

\begin{table*}[t!]
\small{
\begin{center}
\begin{tabular}{@{}c|l|l|c|c|c|c|c|c@{}}
\toprule
\textbf{\begin{tabular}[c]{@{}c@{}}Exp.\\ No.\end{tabular}} &
\multicolumn{1}{c|}{\textbf{Model}} &
\multicolumn{1}{c|}{\textbf{Regularization Configuration}} &
\textbf{\begin{tabular}[c]{@{}c@{}}Train\\ Epoch\end{tabular}} &
\textbf{\begin{tabular}[c]{@{}c@{}}Input\\ Size\end{tabular}} &
\textbf{Top-1} &
\textbf{mCE} &
\textbf{mFR} &
\textbf{Throughput} \\ \midrule
        & EfficientNet B0 \cite{tan2019efficientnet}    & Autoaug                & -             & 224          & 77.3      & 70.7    & -    &  510      \\
        & EfficientNet B1 \cite{tan2019efficientnet}    & Autoaug                & -             & 240          & 79.2      & 65.1    & -    &  352      \\
        & EfficientNet B2 \cite{tan2019efficientnet}    & Autoaug                & -             & 260          & 80.3      & 64.1    & -    &  279      \\
        & EfficientNet B3 \cite{tan2019efficientnet}    & Autoaug                & -             & 300          & 81.7      & 62.9    & -    &  182      \\
        & EfficientNet B4 \cite{tan2019efficientnet}    & Autoaug                & -             & 380          & 83.0      & 60.7    & -    &  95       \\
        & EfficientNet B5 \cite{tan2019efficientnet}    & Autoaug                & -             & 456          & 83.7      & 62.3    & -    &  49      \\ 
        & EfficientNet B6 \cite{tan2019efficientnet}    & Autoaug                & -             & 528          & 84.2      & 60.6    & -    &  28       \\ 
        & EfficientNet B7 \cite{tan2019efficientnet}    & Autoaug                & -             & 600          & 84.5      & 59.4    & -    &  16       \\ \midrule
E0      & R50 (baseline)             &                                           & 120           & 224          & 76.87     & 75.55   & 56.55    & 536       \\ \midrule
E1      & R50\textbf{D}              &                                           & 120           & 224          & 77.37     & 75.73   & 58.17    & 493       \\
E2      & R50D                       & \textbf{LS}                               & 120           & 224          & 78.35     & 74.27   & 54.75    & 493       \\
E3      & R50D                       & LS+\textbf{Mixup}                         & 200           & 224          & 79.10     & 68.19   & 51.24    & 493       \\
E4      & R50D+\textbf{SE}           & LS+Mixup                                  & 200           & 224          & 79.71     & 64.48   & 47.47    & 420       \\
E5      & R50D+SE                    & LS+Mixup+\textbf{DropBlock}               & 270           & 224          & 80.40     & 62.64   & 42.34    & 420       \\
E6      & R50D+\textbf{SK}           & LS+Mixup+DropBlock                        & 270           & 224          & 81.40     & 58.34   & 39.61    & 359       \\
E7      & R50D+SK                    & LS+Mixup+DropBlock+\textbf{KD}            & 270           & 224          & 81.69     & 57.08   & 38.15    & 359       \\
E8      & R50D+SK                    & LS+Mixup+DropBlock+KD                     & \textbf{600}  & 224          & 82.10     & 56.48   & 37.43    & 359       \\
E9      & R50D+\textbf{BL}+SK        & LS+Mixup+DropBlock+KD                     & 600           & \textbf{256} & 82.44     & 55.20   & 37.24    & 359       \\
E10     & R50D+BL+SK+\textbf{AA}     & LS+Mixup+DropBlock+KD                     & 600           & 256          & 82.69     & 54.12   & 36.81    & 312       \\
E11     & R50D+BL+SK+AA              & LS+Mixup+DropBlock+KD+\textbf{Autoaug}    & 600           & 256          & 82.78     & 48.89   & 32.31    & 312       \\ 
E12     & \textbf{R152D}+BL+SK+AA              & LS+Mixup+DropBlock+KD+Autoaug    & 600           & 256          & 84.19     & 43.27   & 29.34    & 143       \\ \bottomrule
\end{tabular}
\end{center}
}

\caption{Ablation study for assembling the network tweaks and regularizations with ResNet-50 on ILSVRC2012 dataset. 
The top-1 accuracy and mCE scores for EfficientNet are borrowed from the official code in~\cite{efficientnet2019offical} and \cite{xie2019adversarial} respectively.
As with other experiments, the inference throughput measurements of EfficientNet were performed on a single Nvidia P40 using official EfficientNet code~\cite{efficientnet2019offical}.
}
\label{tab:assemble}
\end{table*}

Adding ResNet-D to the baseline model improves top-1 accuracy by 0.5\% from 76.87\% to 77.37\% (T1 in Table~\ref{tab:model-tweak}), and adding SK tweaks improves accuracy by 1.46\% from 77.37\% to 78.83\% (T2).
In Table~\ref{tab:channelattention}, We show that the accuracy is increased by 1.62\% when SK is independently applied to ResNet from 76.30\% to 77.92\%. 
Stacking ResNet-D and SK increases the top-1 accuracy gain almost in equal measure to the sum of the performance gains of applying ResNet-D and SK separately.
The results show that the two tweaks can improve performance independently with little effect on each other.
Applying BL to R50D+SK decreases top-1 accuracy from 78.83\% to 78.26\% , but throughput is increased from 359 to 445 (T3).
To achieve higher accuracy by 0.44\% while maintaining throughput similar to that of the R50D+SK, we use $256\times256$ image resolution for inference, whereas we use $224\times224$ image resolution for training (T4). 
Applying AA to the R50D+SK+BL improves top-1 accuracy by 0.12\% from 79.27\% to 79.39\% and decreases throughput by 47 from 359 to 312 (T5).

The ablation study in Table~\ref{tab:assemble} shows the impact of assembling the regularization techniques described in Section \ref{sec:regularization}. The regularization techniques increase accuracy, mCE and mFR altogether, but the performance improvement of mCE and mFR is greater than the improvement of accuracy (E2, 3, 5, 7, and 11).
It can be seen that regularization helps to make CNNs more robust to image distortions.
Adding SE improves top-1 accuracy by 0.61\% and improves mCE by 3.71\% (E4).
We confirm that channel attention is also helpful for robustness to image distortions.
Replacing SE with SK improves top-1 and mCE by 1.0\% and 4.3\% (E6).
In Table~\ref{tab:channelattention}, when SE is changed to SK without regularization, the accuracy increases by 0.5\%.
With regularization, replacing SE with SK nearly doubles the accuracy improvement (E5 and E6).
This means that SK is  more complementary for regularization techniques than SE.

Changing the epochs from 270 to 600 improves performance (E8).
Because data augmentation and regularization are stacked, they have a stronger effect of regularization, so longer epochs seems to yield better generalization performance. 
BL shows a performance improvement not only on top-1, but also on mCE and mFR without inference throughput loss (E9). 
AA also shows higher performance gain in mCE and mFR relative to top-1 (E10), which agrees with AA being used as a network tweak to make the CNN robust for image translations as claimed in \cite{zhang2019shiftinvar}. 

The assembled model of all the techniques described so far has top-1 accuracy of 82.78\%, mCE of 48.89\% and mFR of 32.31\%.
This final model is listed in Table~\ref{tab:assemble} as E11, and we call this model \textbf{Assemble-ResNet-50}.
We also experiment with ResNet-152 for comparison as E12, we call this model \textbf{Assemble-ResNet-152}.

\begin{table*}[h!]
\small{
\begin{center}
\begin{tabular}{l|l|c|cc|ccc}
\toprule
\multicolumn{1}{c|}{\textbf{Model}} &
  \multicolumn{1}{c|}{\textbf{Regularization Configuration}} &
  \textbf{Input Size} &
  \textbf{Top-1} &
  \textbf{mCE} &
  \textbf{FLOPS} &
  \multicolumn{1}{l}{\textbf{Params}} &
  \textbf{Throughput} \\ \midrule
R50 (baseline) &                                                           & 224 & 76.87 & 75.55 & 4.1B & 25.5M & 536  \\ \midrule
R101           &                                                           & 224 & 78.35 & 71.32 & 7.9B & 44.6M & 330 \\ 
R152           &                                                           & 224 & 78.51 & 68.95 & 11.6B & 60.2M & 233 \\ 
R50D+SK+BL           &                                                     & 256 & 79.27 & 67.59 & 5.4B & 41.8M & 359 \\ \midrule
R101           & \begin{tabular}[c]{@{}l@{}}LS+Mixup+DropBlock+KD\end{tabular} & 224 & 81.26  & 57.26  & 7.9B & 44.6M & 330 \\
R152           & \begin{tabular}[c]{@{}l@{}}LS+Mixup+DropBlock+KD\end{tabular} & 224 & 81.96  & 54.99  & 11.6B & 60.2M & 233 \\
R50D+SK+BL     & \begin{tabular}[c]{@{}l@{}}LS+Mixup+DropBlock+KD\end{tabular} & 256 & 82.44 & 55.20 & 5.4B & 41.8M & 359  \\ \midrule
\end{tabular}
\end{center}
}
\caption{Performance comparison among ``ResNet-50D\texttt{+}SK\texttt{+}BL\texttt{+}Regularization'', ``ResNet-101\texttt{+}Regularization'' and ``ResNet-152\texttt{+}Regularization'' on ILSVRC2012 dataset.}
\label{tab:resnet101}
\end{table*}

To further show that the boosted performance of the proposed ResNet-50 is not mainly due to the increase in network parameters, we compared ResNet-50 with network tweaks and regularizations (E9 in Table~\ref{tab:assemble}) to ResNet-101 with regularizations with a similar number of parameters. As shown in Table~\ref{tab:resnet101}, ResNet-50 with network tweaks and regularizations shows approximately 1.2\% better performance in top-1, and 2\% in mCE compared to ResNet101 with regularizations, while having less parameters and FLOPS. Moreover, ResNet-50 with network tweaks and regularizations outperforms ResNet-152 with regularizations which have far larger parameters and FLOPS. These observations prove that the combination of network tweaks and regularizations in ResNet-50 creates a synergistic effect.

\subsection{Ablation Study on MobileNet}
\label{sec:network-tweaks-result-mobile}

\begin{table*}[h]
\footnotesize{
\begin{center}
\begin{tabular}{@{}c|l|l|c|c|c|c|c|c@{}}
\toprule
\textbf{\begin{tabular}[c]{@{}c@{}}Exp.\\ No.\end{tabular}} &
\multicolumn{1}{c|}{\textbf{Model}} &
\multicolumn{1}{c|}{\textbf{Regularization Configuration}} &
\textbf{\begin{tabular}[c]{@{}c@{}}Train\\ Epoch\end{tabular}} &
\textbf{\begin{tabular}[c]{@{}c@{}}Input\\ Size\end{tabular}} &
\textbf{Top-1} &
\textbf{mCE} &
\textbf{mFR} &
\textbf{\begin{tabular}[c]{@{}c@{}}Throughput\\ (FP32/Quantized)\end{tabular}} \\ \midrule
M0      & MobileNet-V1 (baseline)       &                                 & 120           & 224          & 72.59     & 83.85   & 74.29    & 12.98 / 22.49       \\
M1      & SE-MobileNet-V1 (r=16) (baseline)        &                                 & 360           & 224          & 74.28     & 78.86   & 69.98    & 10.33 / 19.71       \\ \midrule
M2      & MobileNet-V1                  & \textbf{LS}                     & 120           & 224          & 72.66     & 83.05   & 71.89    & 12.98 / 22.49       \\
M3      & MobileNet-V1                  & LS+\textbf{Mixup}               & 200           & 224          & 73.54     & 78.56   & 65.78    & 12.98 / 22.49       \\
M4      & MobileNet-V1                  & LS+Mixup+\textbf{KD}            & 200           & 224          & 74.18     & 77.51   & 64.41     & 12.98 / 22.49       \\
M5      & MobileNet-V1                  & LS+Mixup+KD                     & \textbf{360}  & 224          & 74.37     & 76.39   & 65.02    & 12.98 / 22.49       \\ \midrule
M6      & SE-MobileNet-V1 (r=16)        & \textbf{LS}+\textbf{Mixup}+\textbf{KD}   & 360           & 224          & 76.42     & 71.67   & 56.73    & 10.33 / 19.71       \\
M7      & SE-MobileNet-V1 (r=\textbf{2}) & LS+Mixup+KD                     & 360           & 224          & 76.82     & 70.67   & 55.76    & 9.9 / 19.32       \\
M8     & SE-MobileNet-V1 (r=2)         & LS+Mixup+KD+\textbf{DropBlock}  & \textbf{900}  & 224          & 77.30     & 68.12   & 49.99    & 9.9 / 19.32       \\ \bottomrule
\end{tabular}
\end{center}
}
\caption{Ablation study for assembling the network tweaks and regularization with MobileNet on ILSVRC2012 dataset. 
In order to measure throughput, we use the standard TFLite Benchmark Tool.
We measure the floating point and quantized performance using single-threaded large core of Google Pixel 3 with batch size 1.
}
\label{tab:assemble-mobile}
\end{table*}

In this section, the results of applying CNN-related techniques to MobileNet-V1~\cite{howard2017mobilenets} are presented. MobileNet-V1, as its name suggests, is a baseline CNN model for use in mobile edge-devices. To follow the design principle of MobileNet, which prioritizes inference speed, we applied the aforementioned techniques such that the reduction in throughput is minimized. Therefore, among network tweaks, only SE was applied to MobileNet-V1 and boosted the accuracy by 1.69 \% (M0, M1). The top-1 accuracy gain of using SE-MobileNet-V1 together with LS+Mixup+KD was 2.05\% more than that of vanilla MobileNet-V1 with the same regularizations applied (M5, M6). In other words, the synergistic effect of using network tweaks and regularizations is also demonstrated in mobile-oriented models. Based on this, we reduced the reduction ratio $r$ of the SE block from 16 to 2 to maximize synergy between network tweaks and regularization. By doing so, we could improve MobileNet-V1's top-1 accuracy by 1\% with minimal throughput loss. However, unlike ResNet, the top-1 accuracy of SE-MobileNet-V1 decreased when DropBlock was applied. As the network capacity of MobileNet-V1 backbone is smaller than that of ResNet, more training epoch and the adjustment of \(keep\_prob\) hyperparameter (from 1.0-0.9 to 1.0-0.95) are needed for DropBlock regularization to have a sufficient effect in MobileNet-V1 model as that in ResNet (M8).


\subsection{Transfer Learning: FGVC}
\label{sec:transfer-fgvc}

\begin{table*}[h]
\small{
\begin{center}
\begin{tabular}{@{}c|l|c|l|cc@{}}
\toprule
\textbf{\begin{tabular}[c]{@{}c@{}}Exp.\\No.\end{tabular}} &
\multicolumn{1}{c|}{\textbf{Backbone Model}} &
\textbf{\begin{tabular}[c]{@{}c@{}}Backbone\\ Top-1\end{tabular}} &
\multicolumn{1}{c|}{\textbf{Regularization}} &
\textbf{\begin{tabular}[c]{@{}c@{}}Food-101\\ Top-1\end{tabular}} &
\textbf{\begin{tabular}[c]{@{}c@{}}Food-101\\ mCE\end{tabular}} \\ \midrule
F0      & R50 (baseline)                         & 76.87      & -                                         & 86.99     & 61.50     \\ \midrule
F1      & R50\textbf{D}                                   & 77.37      & -                                         & 87.63     & 62.12     \\
F2      & R50D+\textbf{SK}                       & 78.83      & -                                         & 89.77     & 57.20     \\
F3      & R50D+SK+\textbf{BL}                    & 79.27      & -                                         & 90.15     & 57.16     \\ \midrule
F4      & R50D+SK+BL+\textbf{AA}                 & 79.39      & -                                         & 90.37     & 56.66     \\
F5      & R50D+SK+BL+AA                          & 79.39      & \textbf{DropBlock}                        & 91.25     & 53.13     \\
F6      & R50D+SK+BL+AA                          & 79.39      & DropBlock+\textbf{Mixup}                  & 91.64     & 48.53     \\
F7      & R50D+SK+BL+AA                          & 79.39      & DropBlock+Mixup+\textbf{Autoaug}          & 91.85     & 41.73     \\
F8      & R50D+SK+BL+AA                          & 79.39      & DropBlock+Mixup+Autoaug+\textbf{LS}       & 91.76     & 41.40     \\ \midrule
F9      & R50D+SK+BL+AA+\textit{\textbf{REG}}    & 82.78      & -                                         & 90.63     & 53.98     \\
F10     & R50D+SK+BL+AA+\textit{REG}             & 82.78      & \textbf{DropBlock}                                 & 91.62     & 51.01     \\
F11     & R50D+SK+BL+AA+\textit{REG}             & 82.78      & DropBlock+\textbf{Mixup}                  & 92.11     & 45.73     \\
F12     & R50D+SK+BL+AA+\textit{REG}             & 82.78      & DropBlock+Mixup+\textbf{Autoaug}          & 92.21     & 41.69     \\
F13     & R50D+SK+BL+AA+\textit{REG}             & 82.78      & DropBlock+Mixup+Autoaug+\textbf{LS}       & 92.47     & 41.99     \\ \bottomrule
\end{tabular}
\end{center}
}
\caption{Ablation study of transfer learning with the Food-101 dataset.
\textit{REG} means that regularization techniques ``LS+Mixup+DropBlock+KD+Autoaug'' are applied during training backbone.
The Food-101 mCE is not normalized by AlexNet’s errors.
We use the augmentation policy which is found by Autoaug on CIFAR-10 in these experiments \cite{cubuk2018autoaugment}.}
\label{tab:food101}
\end{table*}

\begin{table*}[h]
\small{
\begin{center}
\begin{tabular}{@{}l|c|c|cc@{}}
\toprule
\multicolumn{1}{c|}{\textbf{Dataset}} & 
\textbf{\begin{tabular}[c]{@{}c@{}}The state-of-the-art Models \end{tabular}} &
\textbf{\begin{tabular}[c]{@{}c@{}}ResNet-50 \end{tabular}} &
\textbf{Assemble-ResNet-FGVC-50} \\ \midrule
Food-101           & EfficientNet B7 \cite{tan2019efficientnet} \hspace{3mm} 93.0      & 87.0          & 92.5          \\
CARS196      & EfficientNet B7 \cite{tan2019efficientnet} \hspace{3mm} 94.7      & 89.1          & 94.4          \\
Oxford-Flowers     & EfficientNet B7 \cite{tan2019efficientnet} \hspace{3mm} 98.8      & 96.1          & 98.9         \\
FGVC Aircraft      & EfficientNet B7 \cite{tan2019efficientnet} \hspace{3mm} 92.9      & 78.8          & 92.4          \\
Oxford-IIIT Pets   & AmoebaNet-B \cite{huang2019gpipe} \hspace{5mm}          95.9      & 92.5          & 94.3          \\
 \bottomrule
\end{tabular}
\end{center}
}
\caption{Transfer learning results for FGVC. 
Numbers in the table indicate top-1 accuracy.
}
\label{tab:fgvc}
\end{table*}

In this section, we investigate whether the improvements discussed so far can help with transfer learning.
We first analyzed the contribution of transfer learning for each technique.
An ablation study was performed on the Food-101~\cite{bossard2014food} dataset, which is the largest public fine-grained visual classification (FGVC) dataset. 
The basic experiment setup and hyperparameters that differ from the backbone training are described in supplementary material.


As shown in Table~\ref{tab:food101}, stacking network tweaks and regularization techniques steadily improved both top-1 accuracy and mCE for the transfer learning task on the Food-101 dataset.
In particular, comparing the experiments F4-F8 with experiments F9-F13 (in Table~\ref{tab:food101}) shows the effect of regularization on the backbone.
We use the same network structure in F4-F13, but for F9-F13, they have regularization such as Mixup, DropBlock, KD and Autoaug on the backbone. 
This regularization of the backbone gives performance improvements for top-1 accuracy as expected.
On the other hand, the aspect of mCE performance differed from the top-1 accuracy.
Without regularization during fine-tuning such as in F4 and F9, the backbone with regularization leads to better mCE performance than backbone without regularization.
However, adding regularization during fine-tuning narrows the mCE performance gap (F5-8 and F10-13).
For convenience, we call the final F13 model in Table \ref{tab:food101} as \textbf{Assemble-ResNet-FGVC-50}.


We also evaluated Assemble-ResNet-FGVC-50 in Table \ref{tab:food101} on the following datasets: CARS196 (Stanford Cars)~\cite{krause20133d}, Oxford 102 Flowers~\cite{nilsback2008automated}, FGVC-Aircraft~\cite{maji2013fine}, Oxford-IIIT Pets~\cite{parkhi2012cats} and Food-101~\cite{bossard2014food}. 
The statistics for each dataset are as shown in supplementary material.
Table~\ref{tab:fgvc} shows the transfer learning performance. Compared to EfficientNet~\cite{tan2019efficientnet} and AmoebaNet-B~\cite{huang2019gpipe} which are state-of-the-art models for image classification task, our Assemble-ResNet-FGVC-50 model achieves comparable accuracy with 20x faster inference throughput.

\subsection{Transfer Learning: Image Retrieval}
\label{sec:transfer-ir}

\begin{table}[h!]
\small{
\begin{center}
\begin{tabular}{@{}c|l|l|c@{}}
\toprule
\textbf{\begin{tabular}[c]{@{}c@{}}Exp.\\No.\end{tabular}} & \multicolumn{1}{c|}{\textbf{Backbone}}  & \multicolumn{1}{c|}{\textbf{Regularization}}  & \textbf{Recall@1} \\ \midrule
S0      & R50 (baseline)                    &                                & 82.9      \\ \midrule
S1      & R50\textbf{D}                              &                                & 84.2      \\
S2      & R50D+\textbf{SK}                           &                                & 85.4      \\
S3      & R50D+SK+\textbf{BL}                        &                                & 85.2      \\
S4      & R50D+SK+BL+\textbf{AA}                     &                                & 85.1      \\
S5      & R50D+SK                           & \textbf{DropBlock}                      & 85.9      \\
S6      & R50D+SK                           & DropBlock+\textbf{Autoaug}              & 83.7      \\ \midrule
S7      & R50D+SK + \textbf{\textit{REG}}            &                                & 85.2      \\
S8     & R50D+SK + \textit{REG}             & \textbf{DropBlock}                      & 85.9      \\
S9     & R50D+SK + \textit{REG}             & DropBlock+\textbf{Autoaug}              & 84.0      \\ 
\bottomrule
\end{tabular}
\end{center}
}
\caption{Ablation study of transfer learning with SOP dataset. \textit{REG} means ``LS+Mixup+DropBlock+KD''.}
\label{tab:sop}
\end{table}

We also conducted an ablation study on three public fine-grained image retrieval (IR) datasets: Stanford Online Products (SOP) \cite{song2016deep}, CUB200 \cite{wah2011caltech} and CARS196 \cite{krause20133d}. We borrowed the zero-shot data split protocol from \cite{song2016deep}.

On top of that, cosine-softmax based losses were used for image retrieval.
In this work, we use ArcFace~\cite{deng2019arcface} loss with a margin of 0.3 and use generalized mean-pooling (GeM)~\cite{radenovic2018fine} for a pooling method without performing downsampling at Stage 4 of backbone networks because it has better performance for the image retrieval task. The basic experiment setup and hyperparameters are described in supplementary material.

In the case of SOP, the degree of the effect was examined by an ablation study with the results listed in Table~\ref{tab:sop}.
The particular combinations of network tweaks and regularizations that worked well on the SOP dataset were different from that for FGVC datasets.
Comparing S2-4, we see that BL and AA did not work well on the SOP dataset.
Among the regularizers, DropBlock works well, but Autoaug does not improve the recall at 1 performance (S2 and S5,6).
Nevertheless, in the best configuration, there was a significant performance improvement of 3.0\% compared to the baseline ResNet-50.
The recall at 1 results for image retrieval datasets are reported in Table~\ref{tab:fgir}. 
There is also a significant performance improvement on CUB200 and CARS196 datasets.

\begin{table}[h]
\small{
\begin{center}
\begin{tabular}{@{}l|c|c@{}}
\toprule
\multicolumn{1}{c|}{\textbf{Dataset}}   &
\multicolumn{1}{c|}{\textbf{ResNet-50}} & 
\textbf{\begin{tabular}[c]{@{}c@{}}Assemble-ResNet-IR-50\end{tabular}} \\ \midrule
SOP       & 82.9    & 85.9          \\
CUB200    & 75.9    & 80.3          \\
CARS196   & 92.9    & 96.1          \\ \bottomrule
\end{tabular}
\end{center}
}
\caption{Transfer learning for IR task with our method.
Assemble-ResNet-IR-50 represents the best configuration model for each dataset.
The best configurations for each dataset are specified in supplementary material. Numbers in the table indicate recall@1.
}
\label{tab:fgir}
\end{table}

\section{Conclusion}
\label{sec:conclusion}

In this paper, we show that assembling various techniques for CNNs to single convolutional networks leads to improvements of top-1 accuracy, mCE and mFR on the ILSVRC2012 validation dataset. 
Synergistic effects have been achieved by using a variety of network tweaks and regularization techniques together in a single network. 
Our approach has also improved performance consistently on transfer learning such as FGVC and image retrieval tasks. 
More excitingly, our network is not frozen, but is still evolving, and can be further developed with future research.
We expect that there will be further improvements if we change the vanilla backbone to a more powerful backbone.

{
\small
\bibliographystyle{ieee_fullname}
\bibliography{egbib}
}
\clearpage

\begin{appendices}
\appendix

\section{FLOPS and throughput}
\label{sec:appendix-flops_and_throughput}
We observe in several experiments that FLOPS is not proportional to the inference speed of the actual GPU. FLOPS and throughput for variations of EfficientNet~\cite{tan2019efficientnet} and ResNet~\cite{he2016deep} are described in Table~\ref{tab:flops-and-throughput}. For example, FLOPS of EfficientNet B0 is very small compared to that of ResNet-50, but throughput is rather lower.

\begin{table}[h]
\small{
\begin{center}
\begin{tabular}{@{}l|c|r|c@{}}
\toprule
\multicolumn{1}{c|}{\textbf{Model}} & 
\multicolumn{1}{c|}{\textbf{Resolution}} &
\multicolumn{1}{c|}{\textbf{FLOPS}} &
\multicolumn{1}{c}{\textbf{Throughput}} \\ \midrule
EfficientNet B0       & 224               & 0.39B                              & 510                                     \\
EfficientNet B1        & 240            & 0.70B                              & 352                                     \\
EfficientNet B2        & 260            & 1.0B                               & 279                                     \\
EfficientNet B3        & 300            & 1.8B                               & 182                                     \\
EfficientNet B4      & 380              & 4.2B                               & 95                                      \\
EfficientNet B5     & 456               & 9.9B                               & 49                                      \\
EfficientNet B6     & 528               & 19B                                & 28                                      \\
EfficientNet B7       & 600             & 37B                                & 16                                      \\ \midrule
R50                & 224                & 4.1B                               & 536                                     \\
R50D                & 224               & 4.4B                               & 493                                     \\
R50D+SK         & 224                   & 6.2B                               & 359                                     \\
R50D+SK+BL      & 256                   & 5.4B                               & 359                                     \\
R50D+SK+BL+AA      & 256                & 7.5B                               & 312      
       \\
R152D+SK+BL+AA       & 256               & 20.6B                               & 143                                     \\ \bottomrule
\end{tabular}
\end{center}
}
\caption{FLOPS and throughput for variation of EfficientNet and ResNet. We use the TensorFlow official profiler code to measure FLOPS. EfficientNet’s FLOPS is borrowed from~\cite{tan2019efficientnet}. We measured inference throughput for an Nvidia
P40 single GPU using a batch size of 64.
}
\label{tab:flops-and-throughput}
\end{table}

\section{FGVC Task Configuration}
\label{sec:appendix-fgvc_dataset}
In this section, we will describe experimental configurations for public fine-grained visual classification (FGVC) datasets: Food-101~\cite{bossard2014food}, CARS196~\cite{krause20133d}, Oxford 102 Flowers~\cite{nilsback2008automated}, Oxford-IIIT Pets~\cite{parkhi2012cats} and FGVC-Aircraft~\cite{maji2013fine}.

The basic experimental setup and hyperparameters that differ from the backbone training are described as follows.

\begin{itemize}
  \itemsep0em
  \item Initial learning rate is \(0.01\).
  \item Weight decay is set to 0.001.
  \item Momentum for BN is set to $(\max(1 - 10/s, 0.9))$. 
  \item Keep probability of DropBlock starts at \(0.9\) and decreases linearly to \(0.7\) at the end of training
  \item The training epoch varies for each dataset.
\end{itemize}

We use the same hyperparameters for all datasets for transfer learning except for training epochs. The training epochs for each dataset are described in Table~\ref{tab:fgvc-dataset_configuration}.

\begin{table}[h!]
\small{
\begin{center}
\begin{tabular}{@{}l|r@{}}
\toprule
\textbf{FGVC Dataset} &
\multicolumn{1}{c}{\textbf{Training Epochs}} \\ \midrule  
Food-101         & 100       \\
CARS196    & 1,000        \\
Oxford-Flowers   & 1,000        \\
FGVC Aircraft    & 800        \\
Oxford-IIIT Pets & 1,300        \\
\bottomrule
\end{tabular}
\end{center}
}
\caption{Training configuration of FGVC datasets.}
\label{tab:fgvc-dataset_configuration}
\end{table}



\label{sec:appendix-fgvc_statistic}

\begin{table}[h]
\small{
\begin{center}
\begin{tabular}{@{}l|r|r|c@{}}
\toprule
\textbf{Dataset} &
\multicolumn{1}{c|}{\textbf{Train Size}} & 
\multicolumn{1}{c|}{\textbf{Test Size}} &
\multicolumn{1}{r}{\textbf{\# Classes}} \\ \midrule
Food-101         & 75,750       & 25,250    & 101    \\
CARS196   & 8,144        & 8,041     & 196    \\
Oxford-Flowers   & 2,040        & 6,149     & 102    \\
FGVC Aircraft    & 6,667        & 3,333     & 100    \\
Oxford-IIIT Pets & 3,680        & 3,669     & 37     \\
\bottomrule
\end{tabular}
\end{center}
}
\caption{Statistics of FGVC datasets.}
\label{tab:fgvc-dataset_statistics}
\end{table}

\section{IR Task Configuration}
\label{sec:appendix-fgir_dataset}

In this section, we will describe experimental configurations for three public fine-grained image retrieval (IR) datasets: Stanford Online Products (SOP) \cite{song2016deep}, CUB200 \cite{wah2011caltech} and CARS196 \cite{krause20133d}.
The basic experimental setup and hyperparameters are described as follows.

\begin{itemize}
  \itemsep-0em
  \item Image preprocessing resizes to $224\times224$ without maintaining aspect ratio with probability 0.5 and resizes to $256\times256$ and random crop to $224\times224$ with probability 0.5.
  \item Data augmentation includes random horizontal flip with 0.5 probability.
  \item Momentum for BN is set to \(\max(1 - 10/s, 0.9)\). 
  \item Weight decay is set to 0.0005.
  \item Feature size is set to 1536.
  \item The training epoch, batch size, learning rate decay and assembling configuration vary for each dataset.
\end{itemize}

The different parameter settings for each dataset are described in Table~\ref{tab:fgir-dataset_configuration}.
The best configurations for each dataset are specified.

\begin{table}[h!]
\small{
\begin{center}
\begin{tabular}{@{}l|r|r|r|r@{}}
\toprule
\multicolumn{1}{c|}{\textbf{Dataset}} &
\multicolumn{1}{c|}{\textbf{\begin{tabular}[c]{@{}c@{}}Loss\\ Fuction\end{tabular}}} &
\multicolumn{1}{c|}{\textbf{\begin{tabular}[c]{@{}c@{}}Learning\\ rate\end{tabular}}} &
\multicolumn{1}{c|}{\textbf{\begin{tabular}[c]{@{}c@{}}Batch\\ size\end{tabular}}} &
\multicolumn{1}{c}{\textbf{\begin{tabular}[c]{@{}c@{}}Training\\ Epochs\end{tabular}}} \\ \midrule
SOP      & Arcface   & 0.008     & 128   & 60    \\
CUB200   & Softmax   & 0.001     & 32    & 100   \\
CARS196  & Softmax   & 0.01      & 32    & 100   \\ \bottomrule
\end{tabular}
\end{center}
}
\caption{Different hyperparameter settings for IR tasks.}
\label{tab:fgir-dataset_configuration}
\end{table}

\begin{table}[t]
{
\begin{center}
\begin{tabular}{@{}l|l|l@{}}
\toprule
\multicolumn{1}{c|}{\textbf{Dataset}} &
\multicolumn{1}{c|}{\textbf{Backbone}} &
\multicolumn{1}{c}{\textbf{Regularization}} 
\\ \midrule
SOP      & R50D+SK+\textit{REG} & DropBlock \\
CUB200   & R50D+SK+\textit{REG} & DropBlock \\
CARS196  & R50D+SK+\textit{REG} & DropBlock+LS+Autoaug\\ \bottomrule
\end{tabular}
\end{center}
}
\caption{Model configuration for IR tasks. \textit{REG} means ``LS+Mixup+DropBlock+KD''}
\label{tab:fgir-dataset_tweaks}
\end{table}

\end{appendices}
\clearpage
\end{document}